\documentclass[10pt, a4paper]{article}

\usepackage[final]{lrec2026} 
\usepackage{comment}

\usepackage{multirow}
\usepackage{amsmath}
\usepackage{graphicx}
\usepackage{booktabs}
\usepackage{threeparttable}
\usepackage{hyperref}
\usepackage{float}

\title{ParlaSpeech 3.0: Richly Annotated Spoken Parliamentary Corpora of Croatian, Czech, Polish, and Serbian}

\name{Nikola Ljubešić$^{\dagger\ddagger}$, Peter Rupnik, Ivan Porupski, Taja Kuzman Pungeršek}

\address{Jožef Stefan Institute,
         Ljubljana, Slovenia\\
         $\dagger$ Faculty of Computer and Information Science, University of Ljubljana, Slovenia\\
         $\ddagger$ Institute of Contemporary History, Ljubljana, Slovenia \\
         \{nikola.ljubesic, peter.rupnik, ivan.porupski, taja.kuzman\}@ijs.si\\}

\abstract{
ParlaSpeech is a collection of spoken parliamentary corpora currently spanning four Slavic languages --  Croatian, Czech, Polish and Serbian -- with a total size of more than 6 thousand hours. The corpora were built in an automatic fashion from the ParlaMint transcripts and their corresponding metadata, which were aligned to the speech recordings of each corresponding parliament. In this release of the dataset, each of the corpora has been significantly enriched with several automatic annotation layers. The textual modality of all four corpora has been enriched with linguistic annotations and sentiment predictions. Similarly, their spoken modality has been automatically enriched with occurrences of filled pauses, the most frequent type of disfluency in typical speech. Two languages have been additionally enriched with detailed word- and grapheme-level alignments, and the automatic annotation of the position of primary stress in multisyllabic words. With these enrichments, the usefulness of the corpora has been greatly increased for downstream research across multiple disciplines, which we showcase through an analysis of acoustic correlates of sentiment.  All the corpora are made available for download in JSONL and TextGrid formats, as well as for search through a concordancer.
 \\ \newline  }

\begin{document}

\maketitleabstract

\section{Introduction}


Spoken corpora remain scarce compared to their written counterparts, largely due to the technical, logistical and legal challenges of their creation. Collecting natural speech often requires fieldwork, securing participant consent, and carefully eliciting conversation. Transcription and further annotation also demand significant time and expertise, specialized skills, and deep knowledge of the communicative context. These difficulties are compounded when documenting unwritten languages or non-standard varieties (such as dialects, heritage languages, or L2 speech) which exist only in oral form. Even when available, spoken corpora, especially Slavic ones among European languages, are often small, limited in representativeness, and inconvenient to use, which limits their usefulness for linguistic or phonetic analysis despite their importance for both research and speech technology development~\cite{dobrushina2022spoken}.

The CLARIN Resource Family~\citep{fivser2018clarin} for spoken corpora\footnote{\url{https://www.clarin.eu/resource-families/spoken-corpora}} reveals a highly skewed distribution of resources across European languages. While some, like Finnish and German, are well-resourced with thousands of hours of data, many others (including Croatian, Polish and Serbian) have only a small single corpus recorded. Most languages lack a corpus entirely.
However, emerging corpus platforms, such as those hosted on CLARIN
or TalkBank, have contributed significantly to the accessibility of spoken corpora across multiple languages concurrently. 

Parliamentary proceedings offer an exceptional opportunity for building spoken corpora of various official languages due to the open status of the data, and their public status regarding privacy concerns. While there have been parliamentary corpora constructed in the past, they mostly focused only on the basic alignment between transcripts and recordings, having primarily automatic speech recognition (ASR) use cases in mind.

In this paper, we present the ParlaSpeech collection of spoken corpora~\citep{parlaspeech3},\footnote{\url{http://hdl.handle.net/11356/1833}} currently consisting of recordings from four parliaments, with speeches given in four languages. Unlike other parliamentary spoken corpora, the corpora have been enriched with various linguistic and paralinguistic layers, making the corpora highly useful for downstream research in various disciplines. The comparability of the corpora enables cross-lingual insights, which significantly improves the strength of the  resulting findings.

\section{Related Work}


A number of languages are already represented by parliamentary spoken corpora, e.g., ~\citet{virkkunen2023finnish} list corpora of 10 different languages. However, most of these corpora are primarily aimed at supporting development and improvement of automatic speech recognition (ASR) systems. A slight deviation from simplistic datasets, primarily aimed at ASR, is the Norwegian Parliamentary Speech Corpus~\citep{solberg-ortiz-2022-norwegian} which has its speeches automatically transcribed and manually checked, with non-standardized words explicitly marked and annotated with standardized equivalents.
Another example of a parliamentary spoken corpus being more than just an ASR training dataset is the FT Speech Danish parliamentary corpus~\cite{kirkedal20_interspeech}, which includes normalizations of the official parliamentary transcripts, the lexic being aligned with the Språkbanken lexicon, and a grapheme-to-phoneme conversion. However, none of the available corpora (1) cover multiple languages, enabling cross-lingual studies, and, what is even more, (2) enrich the data on as many layers as is the case with the ParlaSpeech corpora.

Outside the realm of parliamentary corpora, there have been activities in automatic enrichment of spoken corpora, enabling researchers to analyze not just the linguistic content, but also the accompanying prosodic and paralinguistic 
signals such as disfluencies, prosodic variation, speaker-specific characteristics, laughter or gestures~\cite{christodoulides-etal-2014-dismo, nasr2014automatically, dobrushina2022spoken, wu2022using, lemmenmeier2023spoken}. Just recently multi-layer automation approaches with high-quality results started to emerge~\cite{ liao2025nvspeechintegratedscalablepipeline}, while there is a consensus in the community that automatic multilayer annotations of spoken corpora are very much needed~\cite{wieczorkowska2025methodology}.



We organize the remainder of this section by types of enrichment layers applied inside this work.


\textbf{Alignment}. Transcript-aligned corpora are the most common type of corpora. All the parliamentary spoken corpora presented in~\citet{virkkunen2023finnish} have been aligned on the level of instances of medium length, mostly up to a duration of 30 seconds, to streamline development of ASR systems. Most parliamentary corpora are aligned via time-stamped output of ASR, which is a practical solution given that ASR output on the spoken modality is needed to align the spoken modality with the official parliamentary transcripts. However, by now it is almost common knowledge that forced aligners such as the Montreal Forced Aligner (MFA)~\cite{mcauliffe2017montreal} significantly outperform ASR-based time alignments~\cite{rousso2024tradition}.   
Moreover, grapheme- or phoneme-level aligned corpora are rare, such corpus being the British National Corpus~\citep{love2017spoken}.

In this paper we apply an MFA-based word-level and grapheme-level alignment model~\cite{ljubevsic2022parlaspeech} on two out of our four languages, showing significant improvements to the baseline ASR-based word alignment.


\textbf{Primary word stress}. Corpora containing stress labels for vowels or syllables are exceedingly rare, especially automatically annotated.
One such corpus is the ISLE Corpus~\citep{atwell2003isle}, consisting of Italian and German learners’ spoken English (18 hr), that contains word, phone and stressed syllable alignments done manually by experts.
Another such corpus is the Russian National Corpus,
specifically its Accentological corpus (135 Mw). However, audio files are not available and their approach is based on software that accents words using a built-in lexicon, again followed by human expert review and correction~\citep{savchuk2009spoken}.

Recently, a spoken dataset of English disyllabic
words was automatically constructed from three datasets by using CNN-based models of accuracy around 90\%  ~\citep{allouche2025does}.

On our datasets, we apply a recently developed model based on a transformer speech encoder, which shows excellent performance in the two languages processed with word-level accuracy of 99\%~\cite{ljubesic25_stress}.





\textbf{Filled pauses}. Corpora with labeled disfluencies tend to be more available than primary stress corpora. However, these corpora often do not comprise of typical speakers. For example, the FluencyBank Timestamped (FBTS)~\cite{romana2024fluencybank} corpus contains 5.3 hours of annotated English speech from people who stutter, the German KSoF~\citep{bayerl2022ksof} corpus contains over 5500 clips of stuttered speech labeled with six dysfluency types, and the English SEP-28k~\citep{lea2021sep} corpus consists of stuttering speakers with segment-level annotations of filled pauses.



Regarding corpora of typical speakers, the Switchboard corpus~\cite{godfrey1992switchboard} has manually added disfluency markers, including filled pauses.
\citet{zayats2019disfluencies} expand on the corpus, by using an automated approach to identify and add missed disfluency annotations.

In this work, we apply, similar as with the primary word stress layer, a recently developed model based on speech transformers, which achieves performance comparable to that of human annotators in all languages covered in this dataset~\cite{ljubevsic2025_fp}.

\textbf{Sentiment}. Sentiment analysis determines the expressed stance towards a target, and has been a hugely popular task on text~\cite{mao2024sentiment}. Recently multimodal approaches have gained significant traction, which also includes speech~\cite{das2023multimodal}.

Regardless of a significant number of training datasets and benchmarks for sentiment identification on speech and the corresponding transcripts~\cite{kaushik2015automatic,shon2022slue}, large spoken corpora annotated with sentiment information are almost non-existent. One example of such a corpus is the extension of Switchboard with manual sentiment labels~\citep{chen2020large}, covering 140 hours of audio.

\begin{table*}[h!]
    \centering
    \caption{Detailed overview of the ParlaSpeech 3.0 corpora. The table presents the size (in hours and words), speaker distribution, and key annotations (filled pauses, grapheme alignment, and stress labels) for the Croatian, Czech, Polish, and Serbian datasets, covering the specified time ranges.}
    \smallskip
    \renewcommand{\arraystretch}{1.3}
    \setlength{\tabcolsep}{6pt}
    \small
    \begin{tabular}{l|rrrr|r}
        \hline
        \textbf{Size in}             & \textbf{Croatian} & \textbf{Czech} & \textbf{Polish} & \textbf{Serbian} & \textbf{Total} \\
        \hline
        Hours                        & 3\,110            & 1\,218         & 1\,009          & 896              & 6\,233         \\
        Words                        & 24,755,742        & 9,114,266      & 7,515,333       & 7,024,294        & 48,409,635     \\
        Sentences                    & 922,679           & 717,682        & 535,465         & 290,778          & 2,466,604      \\
        Speakers (M/F)               & 348M / 145F       & 431M / 124F    & 610M / 226F     & 398M / 230F      & 1787M / 725F   \\
        Political parties            & 32                & 15             & 10              & 50               & 107            \\
        Filled pauses                & 323,514           & 205,528        & 196,671         & 73,861           & 799,574        \\
        Grapheme-level aligned words & 17,765,585        & --             & --              & 3,138,747        & 20,904,332     \\
        Stress labels                & 11,341,813        & --             & --              & 1,998,020        & 13,339,833     \\
        Time range                   & 2015-2022         & 2013-2023      & 2017-2022       & 2013-2022        & 2013-2023      \\
        \hline
    \end{tabular}
    \label{tab:parlaspeech_v3}
\end{table*}

On the datasets presented in this paper, we exploit a recently developed multilingual text transformer model specifically developed for the parliamentary domain, showing very strong results on two out of our four languages, with comparable results on languages not seen during fine-tuning~\cite{mochtak-etal-2024-parlasent}.

\section{Dataset}

The ParlaSpeech dataset is a derivative of the ParlaMint~\citep{erjavec2025parlamint} corpora, the result of the CLARIN ERIC flagship project, which provided comparable encodings of transcripts of 29 European national and regional parliaments, together with rich speaker metadata. ParlaSpeech extends ParlaMint with recordings from parliaments aligned on the sentence level via an alignment procedure described in~\citet{ljubevsic2022parlaspeech,ljubevsic2024parlaspeech}. The added value of the third version of the corpus is its speech and text enrichments on various levels, described in the next section.

ParlaSpeech 3.0 covers four parliaments: Croatian (HR), Czech (CZ), Polish (PL) and Serbian (RS). The size of each individual corpus and its composition is shown in Table~\ref{tab:parlaspeech_v3}. The dataset consists overall of around 6 thousand hours of speech, a large number of different speakers with good coverage of both genders. In terms of enrichments, all four corpora have rich speaker metadata from ParlaMint \citep{parlamint-repository}, sentiment predictions, linguistic labels and filled pause predictions. At this point, only the Croatian and the Serbian parliament have forced alignment and primary stress predictions.

\begin{figure*}[t]
    \centering
    \includegraphics[width=1\textwidth]{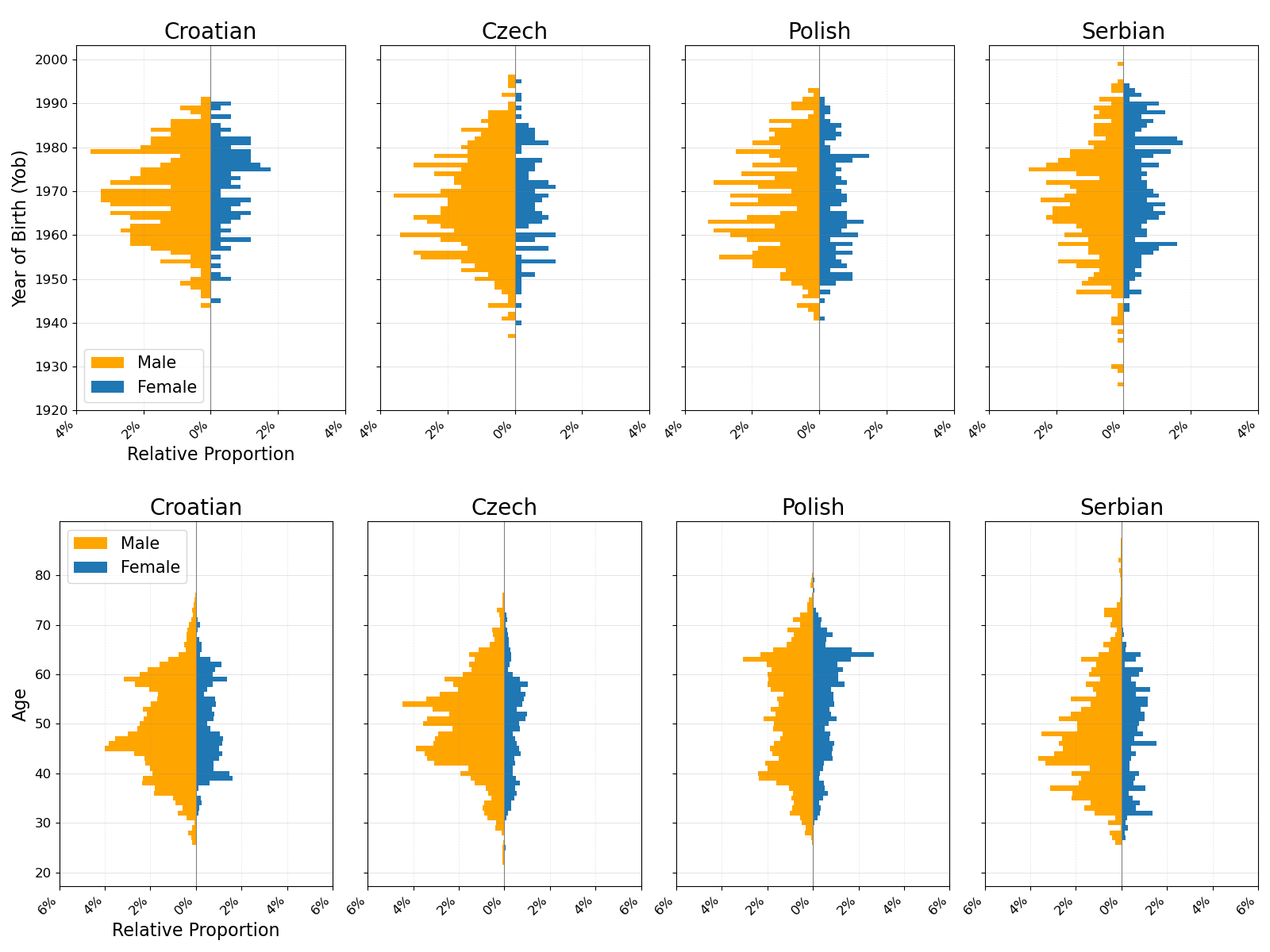}
    \caption{Relative distribution of the number of speakers by year-of-birth and gender across all four parliaments (top), and relative distribution of the number of words spoken by speakers across age, gender, and parliaments (bottom).}
    \label{fig:parlaspeech_age}
\end{figure*}

The content distribution regarding speakers' gender, year of birth and age at the time of the recording are presented in Figure~\ref{fig:parlaspeech_age}. The plots in the first row show the distribution of speakers regarding their year of birth and gender, while the second row shows the distribution of the content (in terms of number of words pronounced) given the gender and age at the time of recording. Each distribution is normalized on the parliament level, with that showing imbalances on the gender and age level inside and across the corpora. We can observe from the plots that there is an expected gender imbalance, which seems to be a little more emphasized on the content level than the speaker level. 
Regarding the age distribution, it seems to be reasonable and expected, with a median age of 50, and infrequent younger speakers.


\section{Dataset Enrichments}

ParlaSpeech 3.0 extends the base ParlaSpeech dataset~\citep{PS-HRv2,PS-CZv1,PS-PLv1,PS-RSv1} with five annotation layers: linguistic annotations (ParlaSpeech-Ling), sentiment (ParlaSpeech-Senti), filled pause detection (ParlaSpeech-Pause), precise word-level and grapheme-level alignments (ParlaSpeech-Align), and primary word stress markers (ParlaSpeech-Stress). Currently, Croatian and Serbian corpora contain all annotation layers, while Czech and Polish include only filled pause, sentiment and linguistic annotations. All the structured enrichments facilitate advanced research in interaction of linguistics, prosody, speech disfluencies, political science, and multimodal parliamentary analysis.

\subsection{ParlaSpeech-Ling}

For linguistic annotation of all four corpora state-of-the art tools  following the Universal Dependencies formalism were used. For Croatian and Serbian the CLASSLA-Stanza tool was applied~\cite{tercon2023classla-stanza}, while for Czech and Polish the Stanza tool was used~\citep{qi2020stanza}. Both tools report very high annotation quality for standard text. 

CLASSLA-Stanza performs for Croatian on level of morphosyntax (XPOS) accuracy of around 94\%, while parsing (LAS) accuracy is around 87\%. For Serbian text, it achieves an XPOS accuracy of about 96\% and LAS accuracy of 90~\% \cite{tercon2023classla-stanza}. Stanza reports for Czech an XPOS accuracy of around 95\% and LAS accuracy of around 89\%, while for Polish XPOS accuracy is around 95\% and LAS accuracy around 92\%\footnote{\url{https://stanfordnlp.github.io/stanza/performance.html}}. Given the varying complexity of the used test data, we can overall expect a morphosyntactic performance of around 95\% accuracy, with a syntactic parsing accuracy of around 90\%.

\subsection{ParlaSpeech-Senti}

For annotating the sentiment layer, the domain-specific ParlaSent model
~\citep{parlasent-model} has been applied on the textual modality.
The model is based on an additionally pre-trained XLM-R-large model on 1.7 billion words of parliamentary proceedings
~\citep{xlm-r-parla}, which was then fine-tuned on 13 thousand instances from English, Bosnian, Croatian, Czech, Serbian and Slovak manually annotated sentences~\citep{11356/1868}, discriminating between six levels of sentiment.
Importantly, ablation experiments on parliamentary test data showed the quality of the automatic annotations to be comparable regardless of whether the test language was present in the fine-tuning data or not~\cite{mochtak-etal-2024-parlasent}. In addition, when comparing this model to a state-of-the-art proprietary LLM, both show comparable performance, with the Spearman correlation coefficient twice as high as for a dictionary approach. The ParlaSent model achieved an average correlation of 0.786, the GPT-4o model of 0.807, while the dictionary approach~\cite{young2012affective} achieved an average correlation of 0.408. For more details see~\citet{mochtak2025parlasent}.

\subsection{ParlaSpeech-Pause}

Filled pauses were annotated with the wav2vecbert2 model\footnote{\url{https://huggingface.co/classla/wav2vecbert2-filledPause}}~\citep{wav2vecbert2-filledPause}, with a 20~ms audio frame classification head fine-tuned on the the ROG dataset for training speech processing technologies~\citep{11356/1992}. Evaluation of the model on parliamentary test data on the four languages covered here shows very high performance, with F1 for Croatian of 0.913, Czech of 0.874, Polish of 0.924, and Serbian of 0.940. What is even more interesting, these F1 metrics touch on human performance measured through inter-annotator agreement. Once the disagreement between human and automatic annotation is compared, the automatic approach actually shows to perform better, with a significantly lower number of false negatives, but with a minor increase in false positives~\cite{ljubevsic2025_fp}.

\subsection{ParlaSpeech-Align}


The ParlaSpeech-Align layer is added via a Montreal Forced Aligner (MFA) model trained on an early version of the Croatian ParlaSpeech corpus~\cite{ljubevsic2022parlaspeech}. To measure the quality of this alignment, it was manually evaluated by an expert phonetician on a random sample of 324 multisyllabic words. For each word, the alignments produced by MFA and ASR (available in all four languages as a byproduct of the speech and text alignment procedure described in~\citet{ljubevsic2024parlaspeech}) were compared with the underlying audio. Deviations of aligned boundaries relative to the actual spoken word were classified as Perfect ($<$30 ms), Close (30--100 ms), Bad ($>$100 ms) or Fail (does not match audio). The thresholds reflect practical usability: shifts $<$30 ms are acceptable for phonetic research, while errors $>$100 ms are considered unreliable for that use. However, even the ``Bad'' alignment should be quite sufficient for retrieval purposes. The results of the comparison of the two word-level alignment approaches are shown in Table~\ref{tab:alignment_eval}.

\begin{table}[H]
    \centering
    \caption{Evaluation of word-level alignments by Montreal Forced Aligner (MFA) and Automatic Speech Recognition (ASR).}
    \smallskip
    \begin{tabular}{lcccc}
        \hline
            & Perfect & Close & Bad & Fail \\
             & $<$30 ms & & $>$100 ms & \\
        \hline
        \smallskip
        MFA & 85.8\% & 8.3\% & 5.3\% & 0.6\% \\
        ASR & 39.5\% & 28.7\% & 30.6\% & 1.2\% \\
        \hline
    \end{tabular}
    \label{tab:alignment_eval}
\end{table}

We can very clearly observe that ASR word alignment is decent, with 68.2\% of words being perfect or close, and 98.8\% at least roughly corresponding to the recording, but still around 30\% of the alignments not being suitable for phonetic research. On the other hand, the MFA output has only 5.9\% of words that are not suitable for phonetic research, with around 85.8\% of alignments considered perfect.
Most ASR errors were caused by silent pauses, speaker noise, or difficulty locating boundaries in fricatives and nasals. MFA performed more consistently, but struggled with disfluencies such as filled pauses, repetitions, and repairs.

\subsection{ParlaSpeech-Stress}

The primary word stress layer has been applied over Croatian and Serbian data, as the underlying enrichment technology~\cite{ljubesic25_stress} currently supports only these two languages. The enrichments have been added on top of the ParlaSpeech-Align layer, i.e., grapheme-level alignments. The model used to identify primary stress is a wav2vecbert2 model~\cite{barrault2023seamless} fine-tuned on manually annotated 10 thousand words from the Croatian ParlaSpeech corpus~\cite{prim_stress_dataset}. The model was evaluated on both Croatian and Serbian parliamentary data, in both cases showing word-level accuracy above 99\%~\cite{ljubesic25_stress}. The primary stress predictions were encoded as raw predictions on 20~ms audio frames, as well as a syllable-level annotation, with the longest raw span prediction assigned to the closest syllable nucleus.

The primary motivation to encode primary stress in Croatian is its high dialectal variability, which results in a varying position of the stress even in official communication~\cite{pletikos2023perception}. Serbian, on the other hand, shares a part of the primary stress tradition in Croatian, but with significantly lower variability. The enrichment of this dataset will enable follow-up research in speaker-level and context-dependent primary stress variation in Croatian, with Serbian as a relevant comparison point.

\section{Dataset Encoding}
To facilitate downstream research, we have encoded the dataset into three different formats and made it available through the CLARIN.SI repository~\citep{parlaspeech3}.

\textbf{JSONL}. The primary format of the corpora is the JSONL (JavaScript Object Notation Lines) file format that contains all the layer data, including speaker metadata.
This format is uniquely suited for computational processing because each line is a self-contained, valid JSON object representing a sentence.
This line-by-line structure allows for highly efficient data streaming and parallel processing of the entire dataset without the need to load massive files into memory. Furthermore, the hierarchical nature of JSON logically organizes the complex layer annotations alongside essential speaker metadata.
This structure provides both standardization for cross-lingual research and the flexibility to accommodate future expansions or additional data layers.
Documentation on the full dataset schema (and more) is available in the online documentation~\footnote{\url{https://clarinsi.github.io/parlaspeech/}}.

\textbf{TextGrid}. The JSONL entries have been converted into individual Praat TextGrid files~\citep{boersma2025praat}, a format particularly suited for phonetic analysis.
These files map phonetic information onto a set of time-aligned, named tiers.
Specifically, the TextGrids include the ParlaSpeech-Align tiers, 
the ParlaSpeech-Stress tier, 
and the ParlaSpeech-Pause tier. 
This format includes the Croatian and the Serbian language for which the enrichments are available. This format makes the dataset immediately convenient and user-friendly for phoneticians and linguists who rely on the Praat software. 
Importantly, TextGrids are easily editable and extensible, enabling researchers to add new tiers and annotations for their own downstream analyses. We hope that researchers will use this opportunity and provide the community with additional data for analysis and modeling purposes.

\textbf{Concordancer}. The corpora are also searchable through the online NoSke concordancer, hosted on the CLARIN.SI  
infrastructure\footnote{\url{https://www.clarin.si/ske/\#open}}.
The main purpose of the concordancers is to allow researchers to easily search and explore the corpus data and metadata using the intuitive NoSke interface.
This approach allows users to search for specific words, lemmas, and part-of-speech tags, or use Corpus Query Language (CQL) for more advanced pattern matching.
Crucially, it is also possible to filter results using text types, allowing researchers to restrict searches by attributes such as gender, age, and sentiment. Each concordance can be also played back, either as a recording of the whole sentence, or as a recording of a chunk up to 6 seconds in length.
For ease of use, we provide a tutorial on how to leverage the full functionality of the concordancers\footnote{\url{https://clarinsi.github.io/parlaspeech/concordancer/concordancer-guide.html}}.

\begin{table*}[h]
    \centering
    \begin{threeparttable}
        \caption{Statistical analysis results on the dependence of acoustic and temporal features on the sentiment of the sentence. Speaker average and instance-level measurements are given.}
        \label{tab:statistical_analysis}
        \renewcommand{\arraystretch}{1.2}
        \small
        \begin{tabular}{@{}l*{3}{c}|l*{3}{c}@{}}
            \toprule
            \multicolumn{4}{c}{\textbf{SPEAKER AVERAGE}} & \multicolumn{4}{|c}{\textbf{INSTANCE-LEVEL}} \\
            \cmidrule(lr){1-4} \cmidrule(lr){5-8}
            \textbf{Feature} & \textbf{p-value} & \textbf{RBC} & \textbf{P(Neg>Pos)} & \textbf{Feature} & \textbf{p-value} & \textbf{RBC} & \textbf{P(Neg>Pos)} \\
            \midrule
            \multicolumn{4}{l}{\textbf{Croatian}} & \multicolumn{4}{l}{\textbf{Croatian}} \\
            \cline{1-4} \cline{5-8}
            \addlinespace[0.1em]
            F0           & \textbf{0.0000} & \phantom{-}0.9937 & 0.9615 & F0           & \textbf{0.0000} & \phantom{-}0.3319 & 0.6186 \\
            Intensity    & \textbf{0.0000} & \phantom{-}0.5735 & 0.7356 & Intensity    & \textbf{0.0000} & \phantom{-}0.0931 & 0.5339 \\
            Jitter       & \textbf{0.0000} & \phantom{-}0.5358 & 0.7115 & Jitter       & \textbf{0.0000} & \phantom{-}0.0692 & 0.5238 \\
            Shimmer      & 0.1507 & -0.1149 & 0.4615 & Shimmer      & 0.8793 & -0.0017 & 0.4994 \\
            Speech rate  & \textbf{0.0000} & \phantom{-}0.4003 & 0.6635 & Speech rate  & \textbf{0.0000} & \phantom{-}0.0486 & 0.5160 \\
            \midrule
            \addlinespace[0.5em]

            \multicolumn{4}{l}{\textbf{Czech}} & \multicolumn{4}{l}{\textbf{Czech}} \\
            \cline{1-4} \cline{5-8}
            \addlinespace[0.1em]
            F0           & \textbf{0.0000} & \phantom{-}0.8783 & 0.8678 & F0           & \textbf{0.0000} & \phantom{-}0.1776 & 0.5565 \\
            Intensity    & \textbf{0.0262} & \phantom{-}0.2329 & 0.5702 & Intensity    & \textbf{0.0005} & \phantom{-}0.0516 & 0.5241 \\
            Jitter       & \textbf{0.0197} & \phantom{-}0.2443 & 0.5785 & Jitter       & 0.4152 & \phantom{-}0.0127 & 0.5033 \\
            Shimmer      & \textbf{0.0026} & \phantom{-}0.3153 & 0.6446 & Shimmer      & \textbf{0.0393} & \phantom{-}0.0309 & 0.5096 \\
            Speech rate  & 0.2556 & -0.1191 & 0.4628 & Speech rate  & 0.1410 & -0.0219 & 0.4908 \\
            \midrule
            \addlinespace[0.5em]

            \multicolumn{4}{l}{\textbf{Polish}} & \multicolumn{4}{l}{\textbf{Polish}} \\
            \cline{1-4} \cline{5-8}
            \addlinespace[0.1em]
            F0           & \textbf{0.0000} & \phantom{-}0.9705 & 0.9427 & F0           & \textbf{0.0000} & \phantom{-}0.3313 & 0.6147 \\
            Intensity    & \textbf{0.0000} & \phantom{-}0.7538 & 0.7834 & Intensity    & \textbf{0.0000} & \phantom{-}0.2036 & 0.5749 \\
            Jitter       & \textbf{0.0001} & \phantom{-}0.3632 & 0.6752 & Jitter       & \textbf{0.0003} & \phantom{-}0.0492 & 0.5187 \\
            Shimmer      & 0.9128 & -0.0101 & 0.5223 & Shimmer      & 0.2994 & \phantom{-}0.0137 & 0.5051 \\
            Speech rate  & \textbf{0.0000} & \phantom{-}0.3985 & 0.6306 & Speech rate  & \textbf{0.0004} & \phantom{-}0.0464 & 0.5127 \\
            \midrule
            \addlinespace[0.5em]

            \multicolumn{4}{l}{\textbf{Serbian}} & \multicolumn{4}{l}{\textbf{Serbian}} \\
            \cline{1-4} \cline{5-8}
            \addlinespace[0.1em]
            F0           & \textbf{0.0000} & \phantom{-}0.9541 & 0.8958 & F0           & \textbf{0.0000} & \phantom{-}0.3104 & 0.5996 \\
            Intensity    & \textbf{0.0000} & \phantom{-}0.6633 & 0.7292 & Intensity    & \textbf{0.0000} & \phantom{-}0.1080 & 0.5350 \\
            Jitter       & \textbf{0.0198} & \phantom{-}0.3844 & 0.5833 & Jitter       & \textbf{0.0082} & \phantom{-}0.0655 & 0.5229 \\
            Shimmer      & \textbf{0.0119} & -0.4133 & 0.3750 & Shimmer      & 0.9757 & -0.0007 & 0.5135 \\
            Speech rate  & \textbf{0.0020} & \phantom{-}0.5034 & 0.6875 & Speech rate  & \textbf{0.0022} & \phantom{-}0.0723 & 0.5271 \\

            \bottomrule
        \end{tabular}
        \begin{tablenotes}
            \scriptsize
            \item \textbf{Statistical Notes:} Both analyses use the Paired Wilcoxon signed-rank test. The statistical value W has been left out for brevity.
            \item \textbf{P = Concordance Probability}; $P = 0.5$ means no difference between sentiments, $P > 0.5$ means negatives tend to be higher than positives.
            \item \textbf{RBC = Rank-Biserial Correlation}.
            A positive RBC value corresponds with an effect towards the Negative sentiment.
            p-values < 0.05 are shown in \textbf{bold}.
        \end{tablenotes}
    \end{threeparttable}
\end{table*}

\section{Use Case}

In this section, we discuss one of the many possible use cases of the ParlaSpeech 3.0 dataset -- investigating the differences in acoustic features between speech with positive and speech with negative sentiment. Does a speaker's pitch, loudness or speech rate vary in different sentiments?

This use case cross-references the sentiment layer with the acoustic layer, and there is a significant number of similar layer interactions that can answer relevant research questions. One example are the filled pause layer and the syntactic parsing layer, which enables inspecting the syntactic role of filled pauses. Another one is the primary word stress layer in Croatian and that of the speaker metadata, investigating the consistency of speakers in using one of the two dominant word stress systems in that language.

\begin{figure*}[h!]
    \centering
    \includegraphics[width=1\textwidth]
    {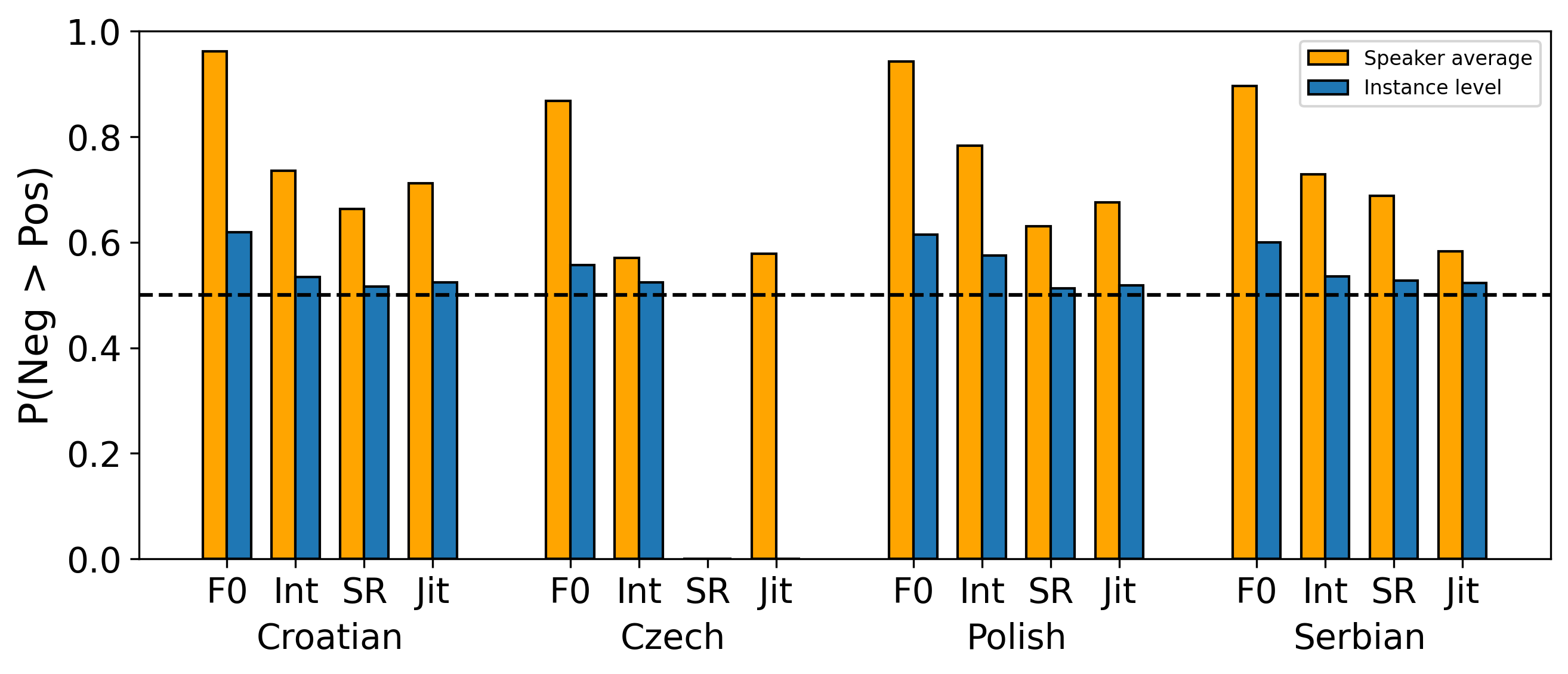}
    \caption{Visualization of the $\mathrm{P}\left(\mathrm{Neg}>\mathrm{Pos}\right)$ effect size on three strong sentiment predictors -- pitch (F0), intensity (Int) and speech rate (SR) -- across the four languages. Speaker average and instance results are shown. Statistically non-significant results (instance-level speech rate in Czech and Polish) are omitted from the plot.}
    
    \label{fig:parlaspeech_probs}
\end{figure*}

\subsection{Data Selection}

To follow upon the question tackled in this section, namely the interaction of sentiment and acoustic features, we first selected only the instances with clear positive or negative predicted sentiment, omitting instances with mixed sentiment. We next filtered the remaining data to include only utterances that were 10 to 40 words long. This length constraint was applied to ensure both a good prediction of sentiment and a consistent sentiment across the instance. 


Finally, we retained only instances of those speakers who had at least 50 positively and 50 negatively labelled instances to allow for good per-speaker estimates.

\subsection{Feature Extraction}

With the instance and speaker selection completed, we moved on to feature selection. To quantify intonation, we extracted the pitch (F0, in Hz) envelope (time series of values). Local jitter, the measure of vocal (F0) perturbation, was also selected as an accompanying feature to intonation.
To quantify loudness, we extracted the intensity (in dB) envelope. We also extracted local shimmer, a measure of variability in the amplitude, representing variation in the intensity signal.

The selected four acoustic features were extracted using Praat and two OpenSMILE extractors (GeMAPS~\cite{eyben2015geneva} and ComParE\_2016~\cite{schuller2016interspeech}). The two OpenSMILE extractors served as a control for Praat, showing highly comparable measurements. In the remainder of the section, only results for Praat extracts are shown.
We also investigated one timing feature, namely speech rate. Speech rate (in syllables/s) was calculated as the number of syllable nuclei (vowels) in a word, divided by the word duration, excluding silent pauses.



Feature extraction was first performed over the entire instance, creating a feature envelope. From the entire envelope, only intervals of actual words were retained, 
skipping over silence and possible noise. These word feature envelopes were then averaged using the median to further counter noise in specific values of the word-level feature envelope. Finally, all word medians, within a single instance, were mean-averaged into one feature value to describe that particular instance.


\subsection{Statistical Analysis}

When comparing the feature values calculated on instances with positive and negative sentiment, we always rely on measurements from the same speaker. There are two reasons for this. The first reason is the high inter-speaker variation of the features in question, esp. the pitch between genders. The second reason is that with this approach, we obtain a paired dataset, which has higher statistical strength than unpaired datasets.

We compare the paired features on two levels: the speaker average level and the instance level. Speaker level analysis averages all positive instances to one value and all negative instances to another, thereby forming one single paired sample per speaker.
Instance-level analysis pairs positive and negative utterances within speakers in a randomized fashion, collecting 50 paired samples per speaker.

For testing statistical significance of the two types of paired samples, the Wilcoxon signed-rank paired test (two-sided) was applied.
For calculating the effect size, we used rank-biserial correlation (RBC) and concordance probability ($\mathrm{P}\left(\mathrm{Neg}>\mathrm{Pos}\right)$).
RBC behaves like any correlation coefficient, ranging from -1 (strong negative correlation), via 0 (no correlation) to 1 (strong positive correlation). Regarding the concordance probability, for the speaker level, the probability states how likely is that the average feature value of a speaker in negative sentiment is higher than the average feature value of the same speaker, but in positive sentiment.
For the instance-level, the probability states how likely is that a random negative-sentiment instance will have a higher feature value than a random positive-sentiment instance of the same speaker.

\subsection{Results}

Statistical analysis revealed consistent patterns in the acoustic correlates of sentiment across all four parliaments, as presented in Table~\ref{tab:statistical_analysis}.
At the speaker-average level, pitch (F0) and intensity emerged as the most robust indicators of sentiment, showing statistically significant differences (p < 0.001) across all parliaments with mostly large effect sizes (RBC = 0.878--0.994 for F0 and RBC = 0.233--0.754 for intensity). At the instance level, both features remained statistically significant across all parliaments, though with low-to-moderate effect sizes (F0: RBC = 0.178--0.332; intensity: RBC = 0.052--0.204).
Jitter showed consistent significance at the speaker-average level across all four parliaments (RBC = 0.244--0.536), always in the positive direction, indicating higher vocal perturbation in negative speech. At the instance level, jitter maintained significance in Croatian, Polish, and Serbian, but not Czech.
Speech rate demonstrated significant effects at the speaker level in Croatian, Polish, and Serbian (RBC = 0.399--0.503), but not in Czech at either level. At the instance level, significance was maintained in the same three languages with small effect sizes (RBC = 0.046--0.072).
In contrast, shimmer showed inconsistent effects: significant in Czech (RBC = +0.315) and Serbian (RBC = -0.413) but in opposite directions, and non-significant in Croatian and Polish, suggesting it reflects language- or speaker-specific variation rather than a systematic sentiment effect. At the instance level, shimmer was non-significant across all languages.

To summarize the findings visually, we represent in Figure~\ref{fig:parlaspeech_probs} the concordance probabilities $\mathrm{P}\left(\mathrm{Neg}>\mathrm{Pos}\right)$ for the four features with consistently positive effects -- pitch (F0), intensity (Int), jitter (Jit), and speech rate (SR) -- across all four parliaments, at both speaker-average and instance levels. The speaker-average probabilities (orange bars) are consistently elevated above the 0.5 random baseline, with F0 showing the strongest effects (0.868--0.962), followed by intensity (0.570--0.783), jitter (0.583--0.712), and speech rate (0.537--0.688). The instance-level probabilities (blue bars) are notably lower across all features and parliaments, clustering closer to the 0.5 border of randomness. We omit Czech speech rate and Czech instance-level jitter results as they were statistically non-significant (p $\ge$ 0.05).

\section{Conclusion}

This paper presented the collection of parliamentary spoken corpora of four Slavic languages, spanning together more than 6 thousand hours in size. The textual modality of the corpora has been automatically enriched with linguistic and sentiment annotations. The spoken modality has been enriched with filled pauses in all four languages, while detailed word- and grapheme-level alignment and subsequent annotation of the primary stress in each multisyllabic word has been applied for two out of the four languages for which the technology is already available and downstream research is most promising.

The corpora are available in various formats by following the FAIR paradigm~\citep{parlaspeech3} -- JSONL for completeness, TextGrid for simple use by phoneticians and other speech scientists, and corpus concordancers for corpus linguists.

We wrapped up the paper with one use case among many potential ones for such a rich dataset -- acoustic and temporal correlates of sentiment. The analysis shows consistent trends in negative sentiment content, when compared to that of positive sentiment, namely strong effects for higher pitch, strong to medium effects for increased intensity, and medium to low effects for increased speech rate and jitter.


In addition to a heavy use of the presented dataset in downstream research, we will continue developing this dataset in two directions: (1) adding additional languages and (2) adding additional annotation layers.


\section{Ethical Considerations and Limitations}

This paper presents a dataset comprising over six thousand hours of human speech in four less-resourced languages. Owing to the public nature of the recordings, we are able to freely enrich and share these data, thereby facilitating further enrichments and analyses by third parties. However, responsible use by downstream users remains essential to ensure the ethical and appropriate application of the dataset.

While the dataset provides an immense opportunity to study spoken communication in the four covered languages, it also entails certain limitations -- including a narrow domain, limited spontaneity in speech production, and imbalances in gender and age representation.

\section{Acknowledgments}

This work was supported in part by the projects ``Spoken Language Resources and Speech Technologies for the Slovenian Language'' (Grant J7-4642), ``Large Language Models for Digital Humanities'' (Grant GC-0002), the research programme ``Language Resources and Technologies for Slovene'' (Grant P6-0411), the Research Infrastructure DARIAH-SI (Grant I0-E007), all funded by the ARIS Slovenian Research and Innovation Agency, and the research project ``Embeddings-based techniques for Media Monitoring Applications'' (L2-50070), co-funded by the Kliping d.o.o. agency.

This research was also supported by LLMs4EU, co-funded by the Digital Europe Programme under GA 101198470. This project is co-funded by the European Union. Views and opinions expressed are however those of the author(s) only and do not necessarily reflect those of the European Union or the European Commission. Neither the European Union nor the granting authority can be held responsible for them.

\section{Bibliographical References}\label{sec:reference}

\bibliographystyle{lrec2026-natbib}
\bibliography{lrec2026-arxiv_incl_lang_res.bib}



\end{document}